\documentclass[10pt,twocolumn,letterpaper]{article}

\usepackage{wacv}
\usepackage{times}
\usepackage{epsfig}
\usepackage{graphicx}
\usepackage{amsmath}
\usepackage{amssymb}
\usepackage{verbatim}

\wacvfinalcopy 

\def\wacvPaperID{****} 

\ifwacvfinal\fi
\begin{document}

\title{Learning to Track Object Position through Occlusion}

\author{Satyaki Chakraborty \\
Carnegie Mellon University\\
{\tt\small schakra1@cs.cmu.edu}
\and
Martial Hebert \\
Carnegie Mellon University\\
{\tt\small hebert@cs.cmu.edu}
}

\maketitle
\ifwacvfinal\thispagestyle{empty}\fi

\begin{abstract}
   Occlusion is one of the most significant challenges encountered by object detectors and trackers. While both object detection and tracking has received a lot of attention in the past, most existing methods in this domain do not target detecting or tracking objects when they are occluded. However, being able to detect or track an object of interest through occlusion has been a long standing challenge for different autonomous tasks. Traditional methods that employ visual object trackers with explicit occlusion modeling experience drift and make several fundamental assumptions about the data. We propose to address this with a `tracking-by-detection` approach that builds upon the success of region based video object detectors. Our video level object detector uses a novel recurrent computational unit at its core that enables long term propagation of object features even under occlusion. Finally, we compare our approach with existing state-of-the-art video object detectors and show that our approach achieves superior results on a dataset of furniture assembly videos collected from the internet, where small objects like screws, nuts, and bolts 
   often get occluded from the camera viewpoint. 
\end{abstract}

\section{Introduction}
        Occlusion often poses a significant 
    challenge in reliably tracking or detecting objects. Current object detection
    tasks\cite{damen2018scaling, deng2009imagenet, everingham2010pascal, lin2014microsoft, ILSVRC15} mostly ignore full occlusion. However, maintaining object position through occlusion can be useful in certain cases and help us better understand video scenes. For instance, in order to encode video demonstrations, it might be useful
    to be able to reliably detect or track small tools that often get occluded 
    from the camera viewpoint. Also, in order to properly predict trajectories of different pedestrians, it might be useful to track them reliably even under occlusion for self driving car applications.
    
    \begin{figure}
        \centering
        \includegraphics[width=0.5\textwidth]{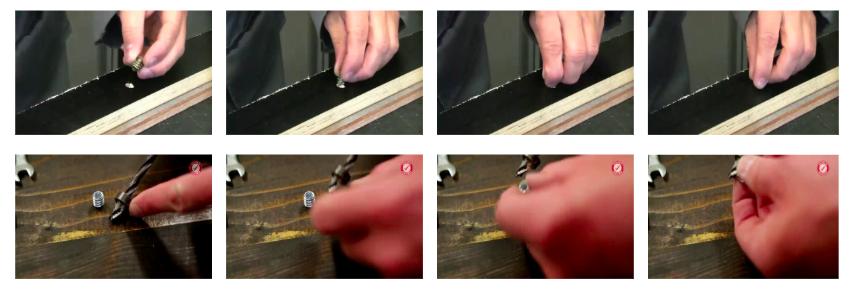}
        \caption{Occlusion observed in furniture assembly demonstrations. Small tools like screws often get occluded from camera viewpoint.}
        \label{fig:occl}
    \end{figure}

        While some older methods model explicit occluder-occludee relationships in visual object trackers to track objects through occlusion, in this paper we plan
    to develop our approach following the more recent \textit{tracking by detection} 
    paradigm. Here, instead  of using visual object trackers to track occluded objects, we first use video object detectors to detect them. The frame level detections obtained from the detector are then linked temporally using association methods like \cite{bewley2016simple, wojke2017simple} to generate tracks. This way we are able to prevent catastrophic failures encountered by visual object trackers where we completely lose track of the object and are unable to track it further even when it has reappeared. Also \textit{tracking by detection} enables us to easily extend our method for multi-object scenarios. In this paper, however, we only concentrate on the detection subproblem, since there has been considerable amount of work done for developing different association methods to solve multi-object tracking.
    
        The detection subproblem is typically addressed by frame level and video level object detectors. Frame level object detectors like Faster RCNN\cite{ren2015faster}, RFCN\cite{dai2016r} can hardly distinguish between the two situations, one where an object is occluded and the other
    where the object is not present. Video object detectors on the other hand, 
    look into a temporal context surrounding the query frame to accumulate features, and then
    reason on top of the accumulated features. This helps the detector to collect
    features of the object of interest from surrounding frames in the video even if the 
    object is occluded in the current frame.
        
        Existing video object detectors \cite{Bertasius_2018_ECCV, feichtenhofer2017detect, han2016seq, kang2018t, kang2016object,
    lee2016multi, xiao2018video, zhu2018towards, zhu2018towardsM, zhu2017flow} are mostly designed to handle partial occlusion, motion blur, unseen viewpoints amongst other issues that frame level detectors are not capable of dealing with. Although we treat full occlusion the same way we treat any of the above mentioned issues, producing bounding boxes when an object is occluded is slightly more challenging mainly because of the following reasons: a) When an object is occluded in a frame, information about the class of the object does not come from that particular frame and hence the architecture needs to heavily rely on the temporal connections to obtain this information. b) The temporal connections usually reason on features belonging to completely different classes of objects to understand occlusion. For example, if a coffee mug is occluded by the hand of the person using the mug, the temporal connections need to combine features of the mug and the human hand and determine that the mug is occluded by the human hand. 
    
    In this research, we aim to solve the occlusion modelling problem in a data driven end-to-end fashion by adding a recurrent computational unit inside region based object detectors following \cite{xiao2018video} to enable propagation of features of the occluded object from both ends of the video. 
    In doing so, we are able to maintain an 
    approximate position of the object through occlusion. We use spatio-temporal memory
    networks from \cite{xiao2018video} as our baseline model and show that our model
    achieves a substantial improvement in terms of raw detection score (mAP) under such
    severe cases of occlusion. Additionally, we show that our method is also able to
    achieve competitive results with the state of the art methods on video object detection datasets like ImageNet VID\cite{imnetvid} which does not deal with complete occlusion of objects.

\section{Related work}
\subsection{Tracking through occlusion}
    Prior work on tracking objects through occlusion with visual object trackers mostly model explicit inter-occlusion relationships between objects in the scene\cite{huang2005tracking}, formulating motion models\cite{pan2007robust}  or use external knowledge\cite{joshi2007synthetic} with visual object trackers. Although such methods are useful when we do not have prior information about the objects that we want to track, visual object trackers suffer from a fundamental issue: once they lose track of the object of interest, they can hardly recover from that. Also, methods that model occlusion explicitly end up making several fundamental assumptions about the data (which includes motion of objects, objects belonging to foreground or background, etc.) which do not make them ideal for real world scenarios. For example, \cite{huang2005tracking} assumes both \textit{occluder} and \textit{occludee} objects belong to the foreground regions and thus learn to model occlusion relationships explicitly by classifying different events of occlusion. Specifically, when an object is getting occluded, they learn to identify it as a foreground region merging event, whereas when an occluded object is becoming visible they learn to identify it as a foreground region splitting event. Thus, by identifying such region merging, splitting and continuation events, their approach is able to track an object through occlusion. Unless the \textit{occluder} objects are known before hand, it is hard to make such methods work in a more general setting.

\subsection{Tracking by detection paradigm}
    The \textit{drift} encountered by traditional visual object trackers can in certain cases lead to irrecoverable failures. Even when the object reappears and is visible, if the appearance model of the tracker has changed considerably, it will not able to track the object of interest further. The situation worsens when there are multiple objects of interest appearing and disappearing through out a video sequence. These issues can be tackled by the recently popular paradigm of tracking by detection. The core idea is that instead of tracking the objects directly across frames with visual object trackers, we use object detectors to produce bounding boxes at every frame. Once we have the bounding boxes from the detector, we use some data association method like \cite{bewley2016simple, wojke2017simple} to link the boxes at different time steps to form tracks. Since there has been considerable amount of work in developing association methods, in this research we only concentrate on the subproblem of robust detection of objects under occlusion. It is worth noting that, a frame level object detector, i.e. an object detector that runs on individual frames of a video is unable to understand occlusion. Hence in order to detect occluded objects it is necessary to use video object detectors which aggregate features from a temporal context surrounding a query frame to produce detections.

\subsection{Frame level object detection}
    Over the last few years, object detection\cite{dai2016r, girshick2015fast, girshick2014rich,  lin2017focal, liu2016ssd, redmon2016you, redmon2017yolo9000, ren2015faster} in static images has received quite a lot of attention. This success can mostly be attributed to very deep convolutional backbones\cite{he2016deep, simonyan2014very}. Earliest of such detectors is a two stage object detector R-CNN\cite{girshick2014rich}, where at first region proposals were computed from the image and then each and every proposal was classified. Later on, computationally lighter versions of the original R-CNN was made possible by leveraging ROI pooling layers\cite{girshick2015fast} and by sharing the convolutional backbone with the region proposal network \cite{ren2015faster}. RFCN\cite{dai2016r} introduces the \textit{position sensitive} ROI pooling layer, and achieves significant speed up compared to \cite{ren2015faster} while achieving competitive detection accuracy. For our application, we mostly build up on Faster RCNN\cite{ren2015faster} and RFCN\cite{dai2016r}, the two most popular region based object detectors.

\subsection{Video level object detection}
    A more recent task in the domain of object detection\cite{Bertasius_2018_ECCV, feichtenhofer2017detect, han2016seq, kang2018t, kang2016object, lee2016multi, xiao2018video, zhu2018towards, zhu2018towardsM, zhu2017flow}, video object detection, has also been given a lot of attention in the past few years. Even though static object detectors can be easily applied to individual frames of a video, there are certain difficult cases where they fail to perform reasonably well. Such difficult cases
    can mostly be attributed to occlusion, motion blur and unseen poses of the objects and these cases make the detection task challenging for the per-frame detector. This is where video object detectors come in. Instead of solely looking at the query frame, video object detectors solve these problems by accumulating features from a temporal context surrounding the given query frame with the hope that rich features of the object can be obtained from at least some of the frames in that context. Most of the recent approaches have concentrated on how to propagate such features efficiently in time.
    
    Although, some of the older methods like \cite{han2016seq, kang2018t, kang2016object, lee2016multi} have heavily relied on exhaustive post processing of detections produced by frame level detectors, more recent approaches\cite{Bertasius_2018_ECCV, feichtenhofer2017detect, xiao2018video, zhu2018towards, zhu2018towardsM, zhu2017flow} in this domain focus on building connections across the convolutional feature maps of the network backbones at different time steps to aggregate features and then reason on such aggregated features. Such methods significantly outperform methods which rely on post-processing of frame level detections. A good number of these methods\cite{Bertasius_2018_ECCV, feichtenhofer2017detect, zhu2017flow} use a short window of frames centered around the frame of interest to accumulate features, while methods like \cite{kang2017object, xiao2018video, zhu2018towardsM} use recurrent computational units like LSTM cell, Spatio-temporal memory module(STMM) and ConvGRU cell\cite{siam2017convolutional} to propagate features in time.

\section{Datasets}
    Most existing datasets for video object detection do not take into consideration objects that undergo full occlusion. Since visual cues of objects are not present when they are fully occluded, existing datasets treat them as if they are not present and hence have no ground-truth annotations for such object instances. This creates the need for datasets with ground-truth annotations for occluded objects. In our datasets, when an object gets fully occluded, we annotate the object based on an approximate guess, which may not be precisely accurate. We also restrict our domain to indoor situations where we try to detect common handheld objects that undergo occlusion (mostly by the human hand). This is because we wanted to avoid certain ambiguous situations in outdoor environments (for example, when a pedestrian gets occluded by a large building, it might be hard to predict the exact location of the pedestrian.)
    
    \begin{figure}[h]
        \centering
        \includegraphics[width=0.5\textwidth]{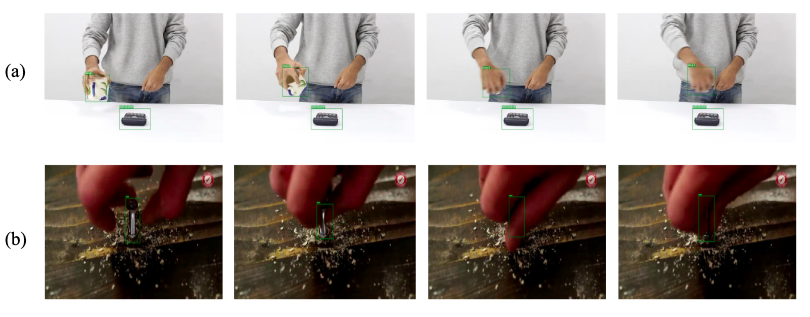}
        \caption{(a) \textit{staged occlusion dataset} with long term occlusion and simpler background settings (b) \textit{furniture assembly dataset} with resemblance to natural occlusion scenes. In both cases, the ground-truth annotations are shown in green.}
        \label{fig:datasets}
    \end{figure}

    \subsection{Staged occlusion dataset}
        The first type of dataset we collect comprises of videos where common hand-held desktop objects like mugs, calculator, notepads, etc. are being manipulated by an individual in front of a simple white background and scenes are captured by a static camera. Objects are manipulated in such a way that they remain occluded by the hand for sufficiently long periods of time from the camera viewpoint while being in motion. Figure \ref{fig:datasets}(a) shows an example scene from this dataset. 
        
        By reducing the noise from external factors like camera movement, scale variation,  background clutter, etc. we are able to narrow down and focus on the effect of long term occlusion. We primarily use this \textit{staged occlusion} dataset to analyze and qualitatively evaluate the impact of occlusion on different methods.
        
    \subsection{Furniture assembly dataset}
        The second type of dataset comprises of different furniture assembly tasks collected from the internet as shown in figure \ref{fig:datasets}(b). This dataset is more representative of occlusions that happen in the natural world. In this case, we try to detect only one class of objects (all small tools like screws, nuts and bolts are grouped into one class). Objects have a lot of scale variation and are often occluded by hands and different tools like hammer and screw-driver. We use this \textit{assembly} dataset for quantitative evaluation by reporting mean average precision (mAP) of different detection methods. \\

        
    \subsection{ImageNet VID}
        Finally, we use the ImageNet VID dataset\cite{imnetvid} to evaluate our method and see how it performs against existing methods for video object detection in datasets which do not target full occlusion although that is not the main goal of this research. This is because in our data driven approach we do not explicitly model occlusion and hope to learn to do it through the temporal connections of the video object detector. Hence ideally our model should be able to adapt to both cases (where occlusion is ignored and where occlusion is annotated) fairly easily.

\section{Methodology}
    \subsection{Building video level architecture}
\begin{figure}[h]
        \centering
        \includegraphics[width=0.45\textwidth]{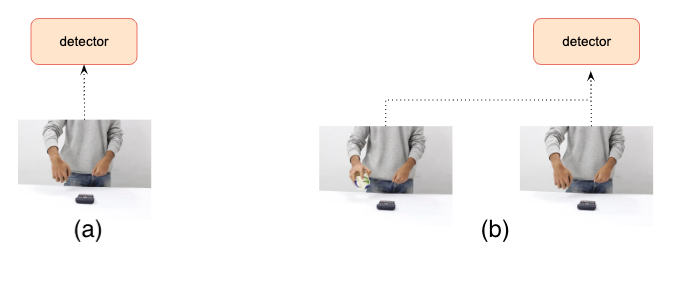}
        \caption{(a) Frame level vs (b) video level object detection. Object is only present on the right hand of the individual while the left hand is empty. The per frame object detector is unable to distinguish between the two cases: one where the object is absent and the other where the object is present but fully occluded. The video level object detector, on the other hand, is able to distinguish between the two cases by exploiting temporal context}
        \label{fig:video_level}
    \end{figure}

    As mentioned earlier, in order to understand occlusion, it is important to utilize the temporal context by building a detector at the video level instead of the frame level. This is better explained in figure \ref{fig:video_level}.
    
    Video object detection methods that use recurrent computational units 
      are not bound by time and thus in theory have the capability of propagating information from the very 
      ends of the video sequence. When an object is occluded, information about the occluded object does not come from the corresponding frames where it remains occluded. Window based approaches like \cite{Bertasius_2018_ECCV, feichtenhofer2017detect, zhu2017flow} can be sub-optimal in this case because the length of the frame window can become a bottleneck. Instead, we build recurrent connections on top of region based object detectors that enable propagation of salient features from both ends of a video sequence. We next explain the architecture details of the video level region based object detector.
      
      \begin{figure}[h]
          \centering
          \includegraphics[scale=0.3]{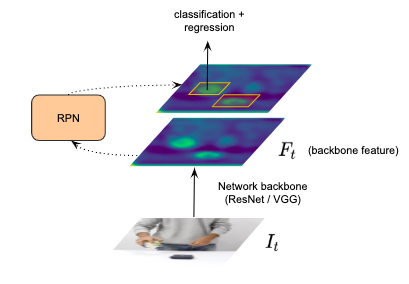}
          \caption{Frame level region based object detector}
          \label{fig:ob_detector}
      \end{figure}
      
      In frame level region based object detectors like Faster RCNN or RFCN, the input frame is passed through a convolutional backbone (typically ResNet or VGG backbones) to obtain a full image backbone feature $F_t$ (subscript denotes time step of frame). A region proposal network (RPN) runs on $F_t$ and produces several ROI crops. Each such ROI crop is then processed further for classification and class specific offset regression. The final offset regression helps in producing slightly tighter or relaxed boxes for better localisation. In order to extend this family of architectures, we first detach the RPN and the ROI specific layers (ROI pooling layers, classification and regression layers) and then add the recurrent connections on top of the backbone feature maps. Following standard convention, we call the output of our RNN cell as memory ($M_t$) since it accumulates features from all past frames of the sequence. We then attach the RPN and the ROI specific layers on top of this memory. Since we are reasoning on top of the accumulated features in the memory and not on the backbone features of a particular time step, we hope to reason on top of features of the object of interest which are propagated to the current memory from when it was last visible in the past. This way we are able to propagate features of the object of interest through occlusion. Our video level object detector is next shown as follows.

      \begin{figure}[h]
          \centering
          \includegraphics[width=0.45\textwidth]{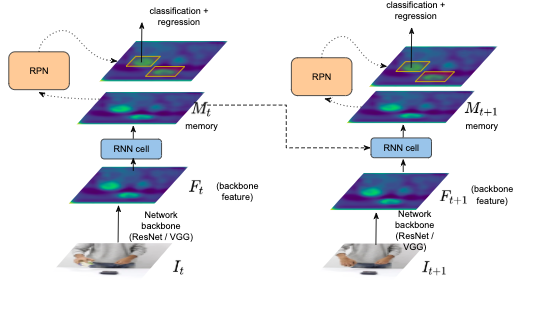}
          \caption{Video level region based object detector}
          \label{fig:vid_ob_detector}
      \end{figure}

\subsection{Choice of recurrent computational unit} 
    ConvGRU, ConvLSTM cells are common choices for the recurrent computational units of ConvRNNs. These cells are inspired from the original GRU and LSTM cells with gating mechanisms, where the dot product layers are replaced by convolutional layers. Recently, another ConvRNN cell called the spatio-temporal memory model\cite{xiao2018video} (abbreviated as STMM) was developed which enables easy transfer of pretrained weights of the frame level detector to the video level detector. In our approach, we build on top of STMM because of its impressive performance on the ImageNet VID dataset and its ability to easily transfer backbone weights pretrained on static image datasets. We observe pretraining the weights of the baseline frame level detector
      to be particularly useful in our case due to the small volume of training videos. We use \textit{Cut, paste and learn}\cite{dwibedi2017cut} to generate synthetic static image datasets to pretrain the weights of the frame level detector.

\subsection{Memory alignment} 
    In practice, a vanilla STMM is unable to align the memory properly. 
    Successive such misalignments end up forming a trail of salient features in the memory, which
    often leads to false positive detections and inaccurate localisation. Xiao and Lee\cite{xiao2018video}
    address this issue by introducing the \textit{MatchTrans} module. They use correlation between the backbone
    features to determine affinity coefficients, which are then used to warp the spatio-temporal
    memory for alignment. 
    
\subsubsection{Effect of explicit memory alignment} 
\begin{figure}[h]
        \centering
        \includegraphics[width=0.5\textwidth]{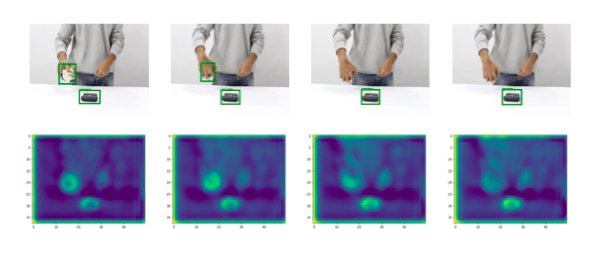}
        \caption{Memory activations of the object held die with time resulting in false negatives. Adjacent time steps shown are 5 frames apart.}
        \label{fig:MatchTransOccl}
    \end{figure}

    While this method of explicit alignment works well for objects that are not completley occluded, we observe that under severe long-term occlusion, correlation based alignment can do away with salient features in the memory of the RNN cell. This happens because, when an object is occluded, backbone feature activations are not always fired for the object (since occluders can often belong to the background class). This in turn results in lower affinity coefficients for the spatial locations where the occluded object exists
    at a given time. Applying \textit{MatchTrans} over successive time steps, results in dying out of the feature activations in memory and can thus result in false negatives as shown in figure \ref{fig:MatchTransOccl}. 
   
    This introduces a trade-off between long-term propagation of features under occlusion and better alignment for accurate localisation. We propose to address this via an alignment learning module, that can act as an alternative to explicit correlation based alignment.

\subsubsection{Learning to the align the memory} 
    \begin{figure}[h]
        \centering
        \includegraphics[width=0.3\textwidth]{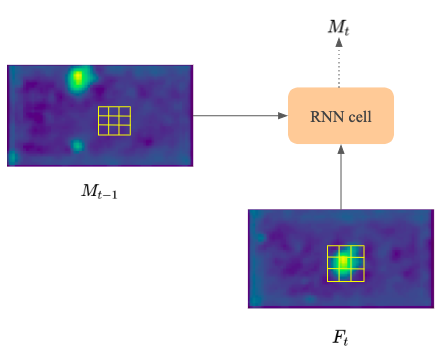}
        \caption{A standard STMM cell using features from the same spatial location of the 
        previous memory and input feature map to form the new memory}
        \label{fig:diff_locs}
    \end{figure}
    Standard implementation of convRNN cells including the STMM cell uses features from the same spatial locations of its inputs, $M_{t-1}$ and $F_t$ to update a cell in its output feature $M_t$. However, unless the objects in a scene are static or moving very slowly, such operations can be problematic, especially since it is a common convention to skip frames from a video to deal with the redundancy of adjacent frames. We believe in order to align memory with standard convolution layers, we should at least ensure large enough receptive fields for the layers of the RNN cell with respect to its input features. A naive implementation of this can be achieved by increasing the kernel size or adding successive 3 by 3 convolution layers. Although simple, such architectures are not memory efficient since adding each convolution layer only increases the receptive field by a finite amount. Hence, the number of parameters, $\Delta P$ needed to be added scales linearly with respect to the increase in the effective receptive field, $\Delta f$ i.e. $\Delta P = O(\Delta f)$. Instead we propose the following  method. 
    
    \begin{figure}[h]
        \centering
        \includegraphics[scale=0.35]{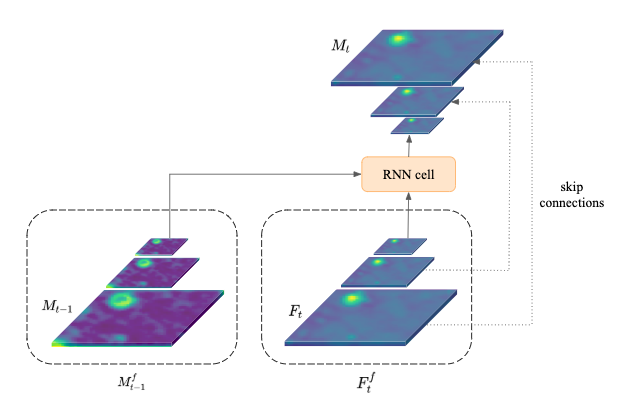}
        \caption{Architecture of alignment learning module}
        \label{fig:architecture}
    \end{figure}
    
    First, we build feature pyramids of the input features of the RNN cell 
    ($M_{t-1}$ and $F_t$). For this step, we use standard 2 x 2 max pooling operation for downsampling, and
    set the number of levels of the pyramid to 3, which gives a good enough balance between memory 
    usage and performance. Feature pyramids of $M_{t-1}$ and $F_t$ are given by $M^f_{t-1}$ and 
    $F^f_t$ where $M^f_{t-1} = \{M_{t-1}, M^{0.5}_{t-1}, M^{0.25}_{t-1}\}$ and 
    $F^f_t = \{F_t, F^{0.5}_t, F^{0.25}_t\}$. Here, numerical superscript denotes scale. We next propagate
    information using only the top most level of the pyramid i.e. using $M^{0.25}_{t-1}$ and $F^{0.25}_t$
    instead of their corresponding full resolution feature maps. This way, we are able to increase the 
    effective receptive field of the layers in the RNN cell without adding more parameters to it. The output of the STMM cell, $M_t$ thus needs to be upsampled to be passed on to other subnetworks of the object detector like the region proposal network, ROI pooling layers etc. In order to upsample the newly updated memory , we use skip connections from the
    backbone feature pyramid $F^f_t$ to combat the information loss due to downsampling
    and to aid the network in better alignment of the memory. Every level of upsampling
    has three fundamental steps. Firstly, we do bilinear upsampling to scale the feature
    maps by 2x followed by an optional zero padding along the width, height or both axes
    to match the spatial resolution of the corresponding feature map from $F^f_t$. This zero padding causes additional misalignment by 1 pixel in the 
    feature space along its corresponding axis. To deal with that, we apply 3 by 3 convolution
    on top of the feature maps accompanied by the skip connections from the backbone feature
    pyramid. The entire architecture of our modified recurrent computational unit is shown in figure \ref{fig:architecture}. This way we also end up adding 
    much fewer parameters to the network. The only parameters that we add are for the 
    skip connections and the number of such parameters linearly increases with the 
    number of levels in the pyramid, $L$. On the other hand, the effective receptive field
    exponentially increases with $L$. Thus, in our model, $\Delta P = O(\log \Delta f)$
    
    \begin{figure}[h]
        \centering
        \includegraphics[width=0.5\textwidth]{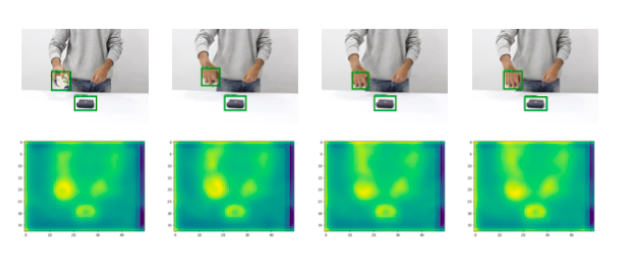}
        \caption{L2 norm of memory and detections for our method. Time increases from left to right. As is apparent, our method of learning to align the memory keeps the features of the object of interest alive in the memory for longer amount of time, resulting in fewer false negatives.}
        \label{fig:lavsmt_occlmem}
    \end{figure}


\section{Results}
    In this section, we evaluate our model both quantitatively and qualitatively on the respective datasets and show the effectiveness of our model in learning the alignment. We use frame level detectors and STMN as baseline models to compare against our method. Unlike \cite{xiao2018video}, we do not take the ensemble of the frame level detector and STMN. Through out all our experiments we only evaluate the single model performance. It is to be noted that all the modules discussed earlier can be plugged into any existing region based object detector with any backbone. For each of the datasets we take different combinations of the convolutional backbone and frame level base network and hence show that our method is invariant of the type of backbone and frame level detector.
    
    \subsection{Experiments on furniture assembly dataset}
    \begin{table}[h]
        \centering
        \begin{tabular}{c  c}
            \hline
            \textbf{Framework hyp} & \textbf{Settings} \\
            \hline
            Base detector & Faster RCNN  \\
            Backbone & ResNet-50 \\
            bptt steps & 5 \\
            Type of RNN & unidirectional \\
            RoI sampling & random \\
            Type of nms & standard \\
            \hline
        \end{tabular}
        \caption{main configurations of video level detector for \textit{Furnuture Assembly} dataset}
        \label{tab:cfg_assm}
    \end{table}
    For both the occlusion datasets, we use Faster RCNN with vgg16 and ResNet-50 backbone as the frame level baseline detector. We train this detector on synthetic datasets generated by \cite{dwibedi2017cut}. 
    Once the Faster RCNN baseline is trained, we add the recurrent connections into the model and fine-tune the entire network in an
    end-to-end fashion. Since we were interested in building online methods, for our case the RNN is uni-directional where information only flows from the beginning to the end. We use stochastic gradient descent with learning rate 1e-3 in the beginning and lower it to 1e-4 as the training loss plateaus. During training, we employ standard left-right flipping for data augmentation and during test time we use standard non-max suppression with an IoU threshold of 0.3. While there are additional techniques to boost the mAP like OHEM\cite{shrivastava2016training} for better ROI sampling, or seq-NMS\cite{han2016seq} for better post processing of raw detections, in this case we do not use them. Under these settings,
    we obtain the following detection scores shown in table \ref{tab:occl_quant}.
    
    \begin{table}[h]
    \begin{center}
    \begin{tabular}{c c c c c}
        \hline
        \textbf{Method} &  \textbf{RNN cell} & \textbf{Alignment} & \textbf{mAP} \\
        \hline
        Frame level & - & - & 0.12\\
        Video level & STMM & - & 0.15\\
        Video level & STMM & MatchTrans & 0.21\\
        Video level & STMM & learned (ours) & \textbf{0.26}\\
        
        \hline
    \end{tabular}
    \caption{mAP of different models on the \textit{Furniture Assembly} dataset with Faster RCNN as base detector.}
    \label{tab:occl_quant}
    \end{center}
    \end{table}
    
\begin{figure*}[h]
        \centering
        \includegraphics[width=1.0\textwidth]{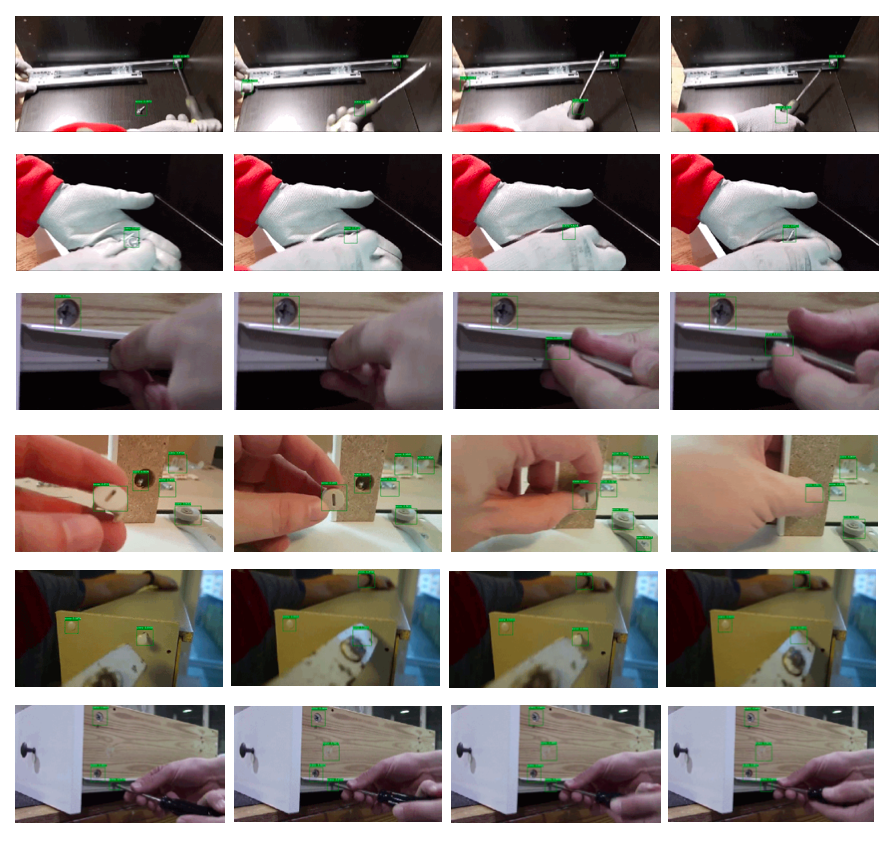}
        \caption{Qualitative results on our \textit{Furniture Assembly dataset}. Each row corresponds to a different sequence. Time increases from left to right. As objects get occluded in the sequence we are able to maintain their approximate positions denoted by the corresponding bounding boxes.}
        \label{fig:qual_assm}
    \end{figure*}

    Unsurprisingly, we observe that our method significantly outperforms the baseline frame level object detector. Also, the detection scores from table \ref{tab:occl_quant} confirm that our method of aligning the features are more suitable under such strong cases of occlusion. We also show qualitative results of our method on this dataset in figure \ref{fig:qual_assm} .

    \subsection{Experiments on ImageNet VID dataset}
\begin{figure}[h]
        \centering
        \includegraphics[width=0.49\textwidth]{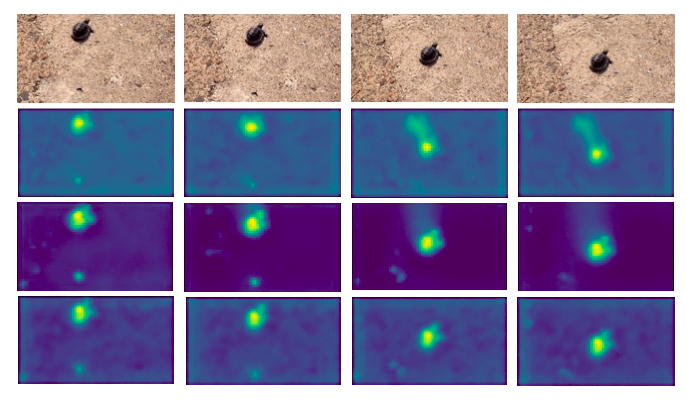}
        \caption{Top row: input sequence (ImageNet VID sequence id ILSVRC2015\_val\_00000000), 2nd row: L2 norm of memory without alignment, 3rd row: L2 norm of memory with MatchTrans, 4th row: L2 norm of memory with learned alignment (our proposed method). Time increases from left to right. Also, the memory activations get cleaner as we move from top to bottom}
        \label{fig:memories_0}
    \end{figure}

\begin{figure}[h]
        \centering
        \includegraphics[width=0.49\textwidth]{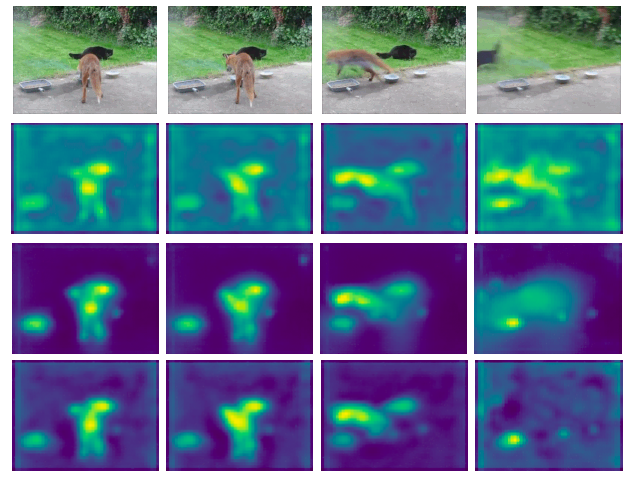}
        \caption{Top row: input sequence, (ImageNet VID sequence id ILSVRC2015\_val\_00037004), 2nd row: L2 norm of memory without alignment, 3rd row: L2 norm of memory with MatchTrans, 4th row: L2 norm of memory with learned alignment (our proposed method). Time increases from left to right. Also, the memory activations get cleaner as we move from top to bottom}
        \label{fig:memories_1}
    \end{figure}

    In our approach, we try to build a data driven end-to-end method for detecting occluded objects in videos and do not plan to model occlusion explicitly. Hence, it is worthwhile to see how our way of learning the alignment compares against explicit alignment with MatchTrans when objects are visible throughout the scene. To do so, we consider the ImageNet VID dataset, a common dataset for benchmarking video object detectors. From figures \ref{fig:memories_0} and \ref{fig:memories_1} we observe that learned alignment gives a relatively better aligned memory when compared to that of \textit{MatchTrans}. 
    
    Further more, we quantitatively evaluate our method's performance on the ImageNet VID dataset to show how our method of video object detection stacks up against current state-of-the-art approaches. In order to make a fair comparison, we make some changes to our method to match the experimental settings of \cite{xiao2018video}. The details are available in table \ref{tab:cfg_imnet}. Our settings differ with that of \cite{xiao2018video} only in two aspects: i) we evaluate single model performance of the video object detector and not performance of the ensemble model with RFCN and ii) during training we unroll the rnn for 4 time steps in stead of 7, because we were unable to fit the latter in a 12 GB Nvidia Titan X GPU. Under these settings, we observe that STMN with \textit{MatchTrans} achieves an mAP of \textbf{0.789} and our STMN with learned alignment achieves an mAP of \textbf{0.796}. Although, we acknowledge that in the case with no occlusion, the improvement is not necessarily statistically significant, we are able to show that our method learns the alignment well enough to act as an alternative to state of the art methods for videos object detection tasks.
    
    \begin{table}[h]
        \centering
        \begin{tabular}{c  c}
            \hline
            \textbf{Framework hyp} & \textbf{Settings} \\
            \hline
            Base detector & RFCN  \\
            Backbone & ResNet-101 \\
            bptt steps & 4 \\
            Type of RNN & bidirectional \\
            RoI sampling & OHEM \\
            Type of nms & seq-NMS \\
            \hline
        \end{tabular}
        \caption{main configurations of video level detector for \textit{ImageNet VID} dataset}
        \label{tab:cfg_imnet}
    \end{table}
    
    
\section{Conclusion and future work}
    In this paper, we present a data driven approach to detecting occluded objects in videos. To the best of our knowledge, prior work on this domain has avoided data driven occlusion reasoning primarily due to lack of available data to train on. Although the advantage of such data driven methods is that we do not need to make any fundamental assumptions about the data, we observe that our method learns some biases for commonly occluding objects that it has seen during training time. As a result, it is unable to generalize to unseen occluder objects at test time. Future work will be concentrated on generalisation across different occluding objects.
    
    Also, without significant volume of training data it is very difficult to make purely data driven occlusion modeling methods work well, and building such datasets with varying levels of occlusion can be laborious. Future work will also target creating synthetic videos and using domain adaptation techniques to address this problem. 

{\small
\bibliographystyle{ieee}
\bibliography{egbib}
}

\end{document}